\documentclass[conference]{IEEEtran}
\IEEEoverridecommandlockouts

\usepackage{cite}
\usepackage{amsmath,amssymb,amsfonts}
\usepackage{algorithmic}
\usepackage{graphicx}
\usepackage{textcomp}
\usepackage{xcolor}
\usepackage{graphicx}
\usepackage{booktabs}
\usepackage[table]{xcolor}
\usepackage{colortbl}
\usepackage{adjustbox}
\usepackage{pgf}
\usepackage{xfp} 

\usepackage{cite}                 
\usepackage{graphicx}             
\usepackage{booktabs}             
\usepackage{array}                
\usepackage{multirow}             
\usepackage{booktabs,makecell,array}
\usepackage{amsmath,amssymb,amsfonts}
\usepackage{mathtools}            

\usepackage{listings}
\usepackage{xcolor}               
\usepackage[T1]{fontenc}     
\usepackage[utf8]{inputenc}  
\usepackage{inconsolata}     
\usepackage{listings}
\lstdefinelanguage{json}{
  basicstyle=\ttfamily\small,
  showstringspaces=false,
  breaklines=true,
  literate=
   *{0}{{{\color{black}0}}}{1}
    {1}{{{\color{black}1}}}{1}
    {2}{{{\color{black}2}}}{1}
    {3}{{{\color{black}3}}}{1}
    {4}{{{\color{black}4}}}{1}
    {5}{{{\color{black}5}}}{1}
    {6}{{{\color{black}6}}}{1}
    {7}{{{\color{black}7}}}{1}
    {8}{{{\color{black}8}}}{1}
    {9}{{{\color{black}9}}}{1}
    {:}{{{\color{black}{:}}}}{1}
    {,}{{{\color{black}{,}}}}{1}
    {\{}{{{\color{black}{\{}}}}{1}
    {\}}{{{\color{black}{\}}}}}{1}
    {[}{{{\color{black}{[}}}}{1}
    {]}{{{\color{black}{]}}}}{1},
  morestring=[b]",
  stringstyle=\color{black},
  comment=[l]{//},
  keywordstyle=\color{black}\bfseries
}
\lstdefinelanguage{bash}{
  alsoletter={-},
  morekeywords={sudo,python,pip,conda,export,CUDA_VISIBLE_DEVICES},
  sensitive=true
}

\usepackage[caption=false,font=footnotesize]{subfig}

\usepackage{microtype}            
\usepackage{balance}              

\usepackage[hidelinks]{hyperref}

\usepackage[nameinlink,capitalize]{cleveref}

\def\BibTeX{{\rm B\kern-.05em{\sc i\kern-.025em b}\kern-.08em
    T\kern-.1667em\lower.7ex\hbox{E}\kern-.125emX}}

\begin{document}

\title{CARScenes: Semantic VLM Dataset for Safe Autonomous Driving}

\author{\IEEEauthorblockN{1\textsuperscript{st} Yuankai He}
\and
\IEEEauthorblockN{2\textsuperscript{nd} Weisong Shi}
}

\maketitle

\begin{abstract}
CARScenes is a frame-level dataset for autonomous driving that enables the training and evaluation of vision–language models (VLMs) for interpretable, scene-level understanding. We annotate 5,192 images drawn from Argoverse1, Cityscapes, KITTI, and nuScenes using a 28-key category/sub-category knowledge base covering environment, road geometry, background-vehicle behavior, ego-vehicle behavior, vulnerable road users, sensor states, and a discrete severity scale (1–10), totaling 350+ leaf attributes. Labels are produced by a GPT-4o–assisted vision–language pipeline with human-in-the-loop verification; we release the exact prompts, post-processing rules, and per-field baseline model performance. CARScenes also provides attribute co-occurrence graphs and JSONL records that support semantic retrieval, dataset triage, and risk-aware scenario mining across sources. To calibrate task difficulty, we include reproducible, non-benchmark baselines—notably a LoRA-tuned Qwen2-VL-2B with deterministic decoding—evaluated via scalar accuracy, micro-averaged F1 for list attributes, and severity MAE/RMSE on a fixed validation split. We publicly release the annotation, and analysis scripts, including graph construction and evaluation scripts, to enable explainable, data-centric workflows for future intelligent vehicles.
\end{abstract}


\section{Introduction}
Recent progress in autonomous driving has been propelled by large-scale datasets that capture diverse environments, multimodal sensors, and richly annotated objects and lanes. Foundational benchmarks such as KITTI \cite{RefWorks:geiger2013vision}, Cityscapes \cite{RefWorks:cordts2016cityscapes}, nuScenes \cite{RefWorks:caesar2020nuscenes}, Argoverse \cite{RefWorks:chang2019argoverse}, Waymo Open \cite{RefWorks:sun2020scalability}, and BDD100K \cite{RefWorks:yu2020bdd100k} have enabled reproducible research across core tasks—2D/3D detection, segmentation, tracking, motion forecasting, and trajectory prediction. Yet, most emphasize low-level perception and spatial reasoning, providing bounding boxes, masks, or trajectories, with limited coverage of high-level scene semantics (e.g., road context, environmental conditions, agent behaviors) that shape human driving decisions.

More recently, as the field increasingly explores vision–language models (VLMs) and language-conditioned interfaces for autonomous driving—spanning language grounding and instruction following \cite{RefWorks:deruyttere2019talk2car} to multi-step reasoning and QA about scenes \cite{RefWorks:sima2024drivelm}—there is a growing need for structured, interpretable scene-level annotations that can both train and evaluate VLMs. General-purpose VLMs (e.g., LLaVA \cite{RefWorks:liu2023visual}, Qwen-VL \cite{RefWorks:wang2024qwen2vl}) demonstrate strong multimodal priors, but lack domain-specific supervision that encodes the semantics and safety cues unique to driving. \textbf{\textit{As a result, we lack per-image, schema-aligned scene-level annotations that are suitable for training and evaluating vision-language models beyond free-text QA or narrow task labels.}}

We introduce CARScenes, a frame-level dataset designed for training and evaluating VLMs on interpretable scene understanding. CARScenes consolidates 5,192 images from Argoverse1, Cityscapes, KITTI, and nuScenes under a 28-key category/sub-category knowledge base (~350+ leaf attributes) that covers environment, road geometry, background-vehicle behavior, ego-vehicle behavior, vulnerable road users, sensor states, and a discrete severity scale (1–10). Labels are produced via a GPT-4o–assisted vision–language pipeline with human-in-the-loop verification; we release the exact prompts, post-processing rules, and per-field model performance for transparency and reuse.

Beyond flat tables, each frame is encoded as an attribute co-occurrence graph, enabling analyses of semantic structure that go beyond task-specific labels. This representation supports semantic retrieval, scenario mining, and content-aware data selection (e.g., identifying redundancy or rare conditions) without additional model training. We further provide a starter, reproducible baseline with a LoRA-tuned Qwen2-VL-2B with deterministic decoding to calibrate task difficulty via scalar accuracy, micro-averaged F1 for list attributes, and severity MAE/RMSE on a fixed validation split. We publicly release the annotation, and analysis scripts, including graph construction and evaluation scripts, to enable explainable, data-centric workflows for future intelligent vehicles. We provide a demo video as a supplementary material and an anonymized interactive demo (ChatGPT GPT, Drive Scene Analyzer): 
\url{https://chatgpt.com/g/g-Rpba6Wp9Q-drive-scene-analyzer}
\section{Related Work}

\begin{table*}[t]
\scriptsize
\centering
\setlength{\tabcolsep}{4pt}
\renewcommand{\arraystretch}{1.2}

\begin{tabular}{
  >{\raggedleft\arraybackslash}m{1.9cm}|
  >{\centering\arraybackslash}m{2.6cm}|
  >{\centering\arraybackslash}m{3.2cm}|
  >{\centering\arraybackslash}m{3.0cm}|
  >{\centering\arraybackslash}m{2.6cm}|
  >{\centering\arraybackslash}m{3.0cm}
}
\toprule
\makecell[r]{\textbf{Dataset}\\\textbf{(year)}} &
\makecell[c]{\textbf{Modality}} &
\makecell[c]{\textbf{Language}\\\textbf{supervision}} &
\makecell[c]{\textbf{Scale}} &
\makecell[c]{\textbf{Provenance}} &
\makecell[c]{\textbf{Release}\\\textbf{type}} \\
\midrule
\makecell[r]{\textbf{CARScenes}\\\textbf{(our work)}} &
\makecell[c]{Single RGB\\frames} &
\makecell[c]{Structured scene KB\\(28 keys; Severity)} &
\makecell[c]{5{,}192 frames\\4{,}692/500 train/val} &
\makecell[c]{Argoverse1, Cityscapes,\\KITTI, nuScenes} &
\makecell[c]{Images + JSONL\\schema+validator+loader} \\
\midrule
\makecell[r]{Talk2Car~\cite{RefWorks:deruyttere2019talk2car}\\(2019)} &
\makecell[c]{Front camera\\(nuScenes)} &
\makecell[c]{Commands with bbox\\(grounding)} &
\makecell[c]{8{,}349/1{,}163/\\2{,}447 cmds} &
\makecell[c]{nuScenes} &
\makecell[c]{Text + boxes\\on nuScenes} \\
\midrule
\makecell[r]{nuScenes-QA~\cite{RefWorks:qian2024nuscenesqa}\\(AAAI'24)} &
\makecell[c]{Multi-modal,\\multi-frame} &
\makecell[c]{Programmatic VQA\\pairs} &
\makecell[c]{$\sim$460K QAs\\over $\sim$34K scenes} &
\makecell[c]{nuScenes} &
\makecell[c]{Benchmark\\code + data} \\
\midrule
\makecell[r]{NuInstruct~\cite{RefWorks:ding2024holistic}\\(CVPR'24)} &
\makecell[c]{Multi-view\\video} &
\makecell[c]{Instruction--response\\(17 subtasks)} &
\makecell[c]{91K\\video--QA pairs} &
\makecell[c]{nuScenes\\(multi-view)} &
\makecell[c]{Annotations\\+ loaders} \\
\midrule
\makecell[r]{DriveLM-Data~\cite{RefWorks:sima2024drivelm}\\(ECCV'24)} &
\makecell[c]{Multi-view images/\\video} &
\makecell[c]{Graph-structured QA\\(P3: perc/pred/plan)} &
\makecell[c]{n/a} &
\makecell[c]{nuScenes\\+ CARLA} &
\makecell[c]{Dataset\\+ eval server} \\
\midrule
\makecell[r]{BDD-X~\cite{RefWorks:kim2018textual}\\(2018)} &
\makecell[c]{Video} &
\makecell[c]{Action descriptions\\+ explanations} &
\makecell[c]{6{,}970 videos; 77 h\\$\sim$26K activities} &
\makecell[c]{BDD} &
\makecell[c]{Explanations\\dataset} \\
\bottomrule
\end{tabular}
\vspace{0.1cm}
\caption{Related driving VLM datasets compared to \textbf{CARScenes}.}
\label{tab:related_datasets}
\end{table*}

\textit{Autonomous-driving datasets.} KITTI, Cityscapes, BDD100K, Argoverse, nuScenes, Lyft Level 5, and Waymo Open provide multimodal data with object/pixel-aligned labels for detection, tracking, and mapping. These corpora include limited, coarse scene metadata and are not designed for schema-consistent scene-level semantics. CARScenes complements them by releasing images + JSONL scene annotations for interpretable analysis rather than leaderboard evaluation.

\textit{Vision–language for driving.} BDD-X, Talk2Car, DriveLM/DriveLM-Data, NuInstruct, and nuScenes-QA pair scenes with free-form text, QA, or graph-QA signals. While powerful, supervision and vocabularies are heterogeneous. CARScenes instead offers a versioned, fixed schema (scene attributes incl. severity) enabling consistent querying and symbolic analysis.

\textit{Scene graphs \& risk corpora.} Visual Genome~\cite{RefWorks:krishna2017visual}, RefCOCO~\cite{RefWorks:chen2025revisiting}, and GQA~\cite{RefWorks:hudson2019gqa} probe general-image relations/referring or VQA~\cite{RefWorks:antol2015vqa}, while Scenic~\cite{RefWorks:fremont2019scenic}, RiskBench~\cite{RefWorks:kung2024riskbench}, DReyeVE~\cite{RefWorks:alletto2016dr}, and DeepScenario~\cite{RefWorks:lu2023deepscenario} center on simulators, trajectories, or attention—none provide a per-image, driving-specific semantic schema (signals, lane/lane-use, ego context) grounded directly in real images. As a result, they don’t support standardized, per-key evaluation of scene-level attributes or lightweight cross-dataset curation. CARScenes addresses this by offering single-frame, schema-aligned annotations for environment, traffic control, road geometry, actors, and a coarse difficulty/severity proxy, with a consistent evaluation protocol that complements these efforts.
\section{Dataset Overview}
\label{sec:dataset}

\subsection{Sources and Scope}
CARScenes augments existing autonomous-driving datasets with \emph{scene-level} annotations and a discrete severity score (1--10). We aggregate \textbf{5,192} images from \textbf{Argoverse1}, \textbf{Cityscapes}, \textbf{KITTI}, and \textbf{nuScenes}, chosen for complementary coverage of geography, weather, and urban form. We release \emph{annotations and tools only}; original imagery remains under the licenses of the source datasets.

\subsection{Schema and Knowledge Base}
\label{sec:schema}

\textbf{Design goals.} CARScenes targets scene-level, interpretable supervision that is (i) \emph{human-readable} and stable across datasets, (ii) \emph{VLM-ready} for training and evaluation, and (iii) \emph{schema-faithful} with explicit enumerations and constraints.

\subsection{28-Key Category/Sub-Category KB (350{+} Attributes)}
\begin{table}[t]
\scriptsize
\centering
\setlength{\tabcolsep}{4pt}
\renewcommand{\arraystretch}{1.15}

\begin{tabular}{>{\raggedleft\arraybackslash}m{.33\linewidth}|>{\centering\arraybackslash}m{.61\linewidth}}
\toprule
\textbf{Group} & \textbf{Fields (subset)} \\
\midrule
\makecell[r]{Environment} &
\makecell[c]{TimeOfDay \\ Weather \\ Visibility.General \\ Visibility.SpecificImpairments} \\
\midrule
\makecell[r]{Road Geometry \& \\ Infrastructure} &
\makecell[c]{RoadType \\ LaneInformation.NumberOfLanes \\ LaneInformation.SpecialLanes \\ LaneInformation.LaneMarkings \\ IntersectionType} \\
\midrule
\makecell[r]{Traffic Control} &
\makecell[c]{TrafficSigns.Types \\ TrafficSigns.TrafficSignsVisibility \\ TrafficLights.TrafficLightState} \\
\midrule
\makecell[r]{Background\hspace{1pt}-Vehicle \\ Behavior} &
\makecell[c]{Vehicles.TotalNumber \\ Vehicles.VehicleTypes \\ Vehicles.InMotion \\ Vehicles.States} \\
\midrule
\makecell[r]{Ego\hspace{1pt}-Vehicle \\ Behavior} &
\makecell[c]{EgoDirection \\ EgoManeuver} \\
\midrule
\makecell[r]{Vulnerable Road Users \\ (VRUs)} &
\makecell[c]{Pedestrians \\ Cyclists} \\
\midrule
\makecell[r]{Sensor States} &
\makecell[c]{CameraCondition } \\
\bottomrule
\end{tabular}
\vspace{0.1cm}
\caption{Overview of the 28-key category/sub-category knowledge base (KB). Full schema includes 350{+} leaf attributes.}
\label{tab:kb_overview}
\end{table}

Each frame also carries a bounded integer \texttt{Severity} on a \textbf{1--10} discrete scale.

\paragraph{Field types and naming.} Keys are either \emph{scalar categorical} (single enum), \emph{list-valued} (set of attributes), or \emph{bounded-integer} (\texttt{Severity}). We use \emph{dot-delimited} names (e.g., \texttt{TrafficSigns.Types}) and canonical enumerations (e.g., \texttt{Urban}, \texttt{Highway}, \texttt{Night}).

\paragraph{Severity definition.} We use severity as an image-only, ordinal difficulty/criticality proxy (1–10) determined from visible cues that plausibly elevate the burden and stakes of scene understanding. The severity score considers (i) road participant complexity (counts, proximity, occlusions), (ii) road/geometry complexity (intersections, merges, lane drops, work zones), (iii) traffic control clarity (visibility/ambiguity of lights, signs, markings), (iv) visibility/conditions (rain, fog, low light, glare, lens artifacts), (v) anomalies/events (stopped vehicles, emergency scenes, debris), and (vi) hazard potential from static cues (e.g., a pedestrian could plausibly emerge from behind a parked vehicle near a crosswalk or on-street parking spaces). Asymmetric consequences are incorporated during labeling: mistakes that would be consequential in context (e.g., treating a red light as green at a signalized junction) increase the assigned severity more than the converse.

\paragraph{Missingness policy.} For list-valued keys, images without that annotation are \emph{excluded} from list micro-averages; for scalar keys, unannotated fields are omitted from accuracy aggregates. We never conflate “absent” with “unobserved.”

\subsection{Canonical Record and File Format}
All annotations are distributed as \textbf{JSONL} (one record per frame) with exact key names/values validated against a machine-readable schema. A typical record:

\begin{lstlisting}[language=json, basicstyle=\ttfamily\small]
{
  "Scene": "Industrial",
  "TimeOfDay": "Daytime",
  "Weather": "Sunny",
  "RoadConditions": "Dry",
  "LaneInformation": {
    "NumberOfLanes": "Two",
    "LaneMarkings": "LaneVisible",
    "SpecialLanes": ["NoSpecialLanes"]
  },
  "TrafficSigns": {
    "TrafficSignsTypes": ["NoTrafficSigns"],
    "TrafficSignsVisibility": "SignNotVisible"
  },
  "Vehicles": {
    "TotalNumber": "MultipleVehicles",
    "InMotion": ["False"],
    "States": ["Parked"],
    "VehicleTypes": ["Pickup Truck", "SUV", "Sedan"]
  },
  "Pedestrians": [
    "NoPed"
  ],
  "Directionality": "Two-Way",
  "Ego-Vehicle": {
    "EgoDirection": "EgoForward",
    "EgoManeuver": "EgoMoving"
  },
  "Visibility": {
    "General": "Good",
    "SpecificImpairments": ["NoImpairments"]
  },
  "CameraCondition": "Clear",
  "Severity": 2
}
\end{lstlisting}

\subsection{Annotation Pipeline and Quality}
\label{sec:annotation}

\textbf{Objective.} Produce schema-faithful labels that are deterministic, auditable, and suitable for training/evaluating VLMs.

\subsection{VLM-Assisted Labeling}
We prompt a vision–language model with the exact KB and require a single JSON object with a paragraph describing the scene. Decoding is \emph{deterministic} (temperature~0.0, top-$p$~1.0, $n{=}1$). The prompt embeds a \texttt{schema\_version} and instructs the model to select only from allowed values.

\subsection{Post-Processing and Canonicalization}
A deterministic validator enforces:
\begin{enumerate}
  \item \textbf{Schema checks:} type/enumeration validation, dot-delimited keys, canonical case/spelling.
  \item \textbf{Synonym mapping:} controlled rewrites from common paraphrases to KB enums.
  \item \textbf{Cross-field constraints:} e.g., presence-before-state for \texttt{TrafficLights.State}, mutual exclusions for \texttt{TimeOfDay}, uniqueness for list fields.
  \item \textbf{Normalization:} sorted, de-duplicated lists for hash-stable records.
\end{enumerate}
Violations are logged and routed to human review.

\subsection{Human-in-the-Loop Verification}
We use (i) \emph{triggered review} for any record that fails validation or contains rare/ambiguous combinations and (ii) a \emph{stratified random audit} by source, weather, and \texttt{Severity}. Reviewers edit via a schema-bound GUI (dropdowns only); edits are logged and become the final labels.

\subsection{Reliability and Agreement}
We quantify label reliability with field-appropriate metrics:
\begin{itemize}
  \item \textbf{Scalar fields:} per-key exact-match accuracy and a macro average across scalar keys.
  \item \textbf{List-valued fields:} micro-averaged precision/recall/F1 over leaf attributes (denominators exclude images without that field).
  \item \textbf{Severity (1--10):} 10-way classification accuracy and regression MAE/RMSE.
\end{itemize}

\section{Dataset Statistics and Structure}
\label{sec:stats}

This section quantifies the scale and semantic coverage of \textbf{CARScenes} and documents how frames are sampled from source datasets. Our goal is to characterize attribute diversity and severity (1--10) distributions in a way that is directly useful for training and evaluating VLMs.

\subsection{Data Sources and Splits}

\textbf{Sources.} CARScenes comprises annotated frames from four widely used autonomous-driving datasets: \emph{Argoverse1}, \emph{KITTI}, \emph{Cityscapes}, and \emph{nuScenes}. These sources jointly cover varied geographies, road types, traffic mixes, and environmental conditions.

\textbf{Sampling protocol.} We uniformly subsample each dataset’s \emph{front-facing RGB} stream at fixed temporal intervals to reduce near-duplicate frames while preserving transitions between scenarios (e.g., intersection approaches, merges). All selected frames are annotated under the same 28-key knowledge base.

CARScenes includes \textbf{5,192} frames sampled from Argoverse1, Cityscapes, KITTI, and nuScenes. We provide a training split of \textbf{4,692} images and a fixed validation split of \textbf{500} images. Users may re-partition by source/city/condition using provided scripts.

\begin{figure*}[t]
    \centering
    \includegraphics[width=\linewidth]{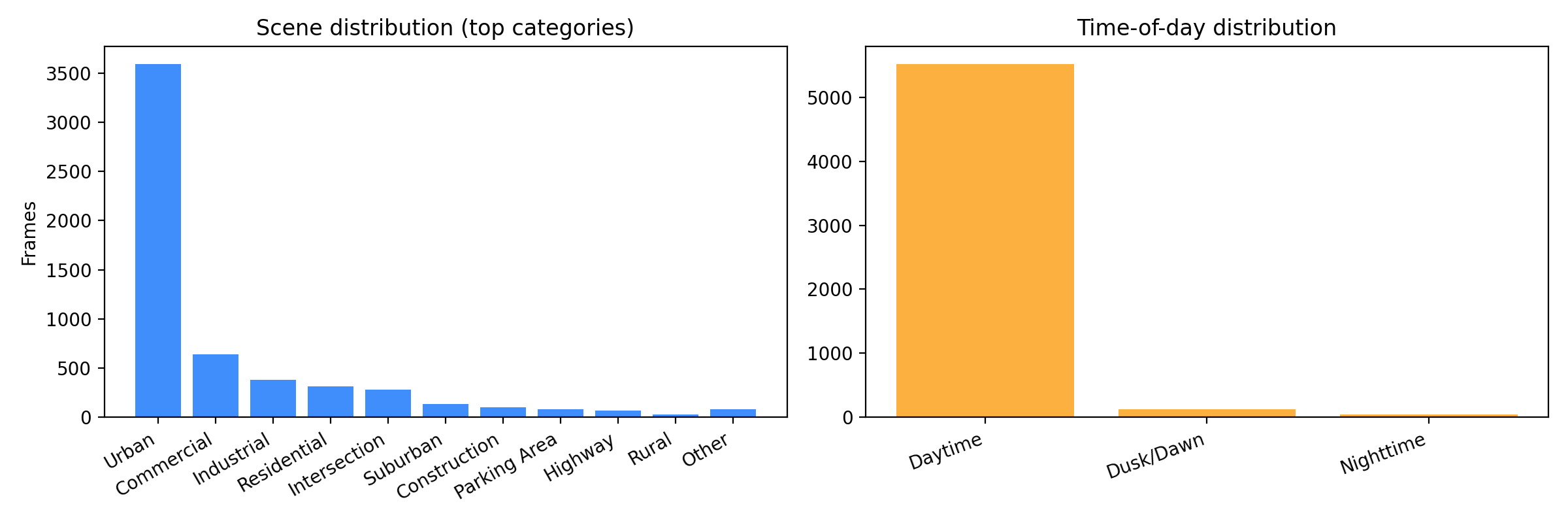}
    \caption{Dataset attribute distribution (scenes and time-of-day).}
    \label{fig:dataset-attributes}
\end{figure*}

\begin{figure}[t]
    \centering
    \includegraphics[width=\linewidth]{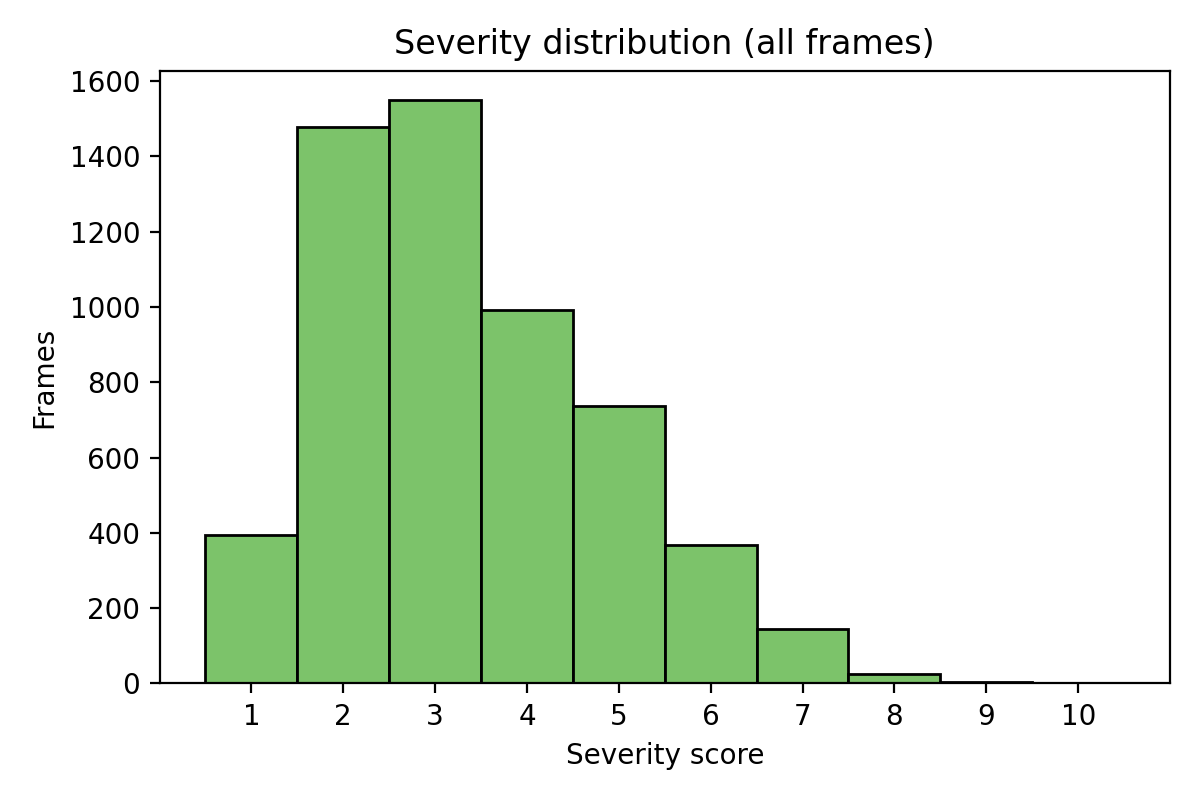}
    \caption{Severity histogram over all frames.}
    \label{fig:dataset-severity}
\end{figure}

All multimodal supervision is prepared with \texttt{utils/qwen\_custom\_dataset.py} using the prompt template defined in \texttt{configs/qwen\_dataset.yaml}. The converter pairs each RGB frame with its structured annotation, generating a JSONL record that contains (i) a natural-language analyst summary and (ii) the raw \texttt{Overall.json} label appended verbatim. The resulting training corpus contains $4\,692$ samples drawn from four driving benchmarks (Table~\ref{tab:split}) while validation uses a fixed, held-out file at \texttt{val\_dataset\_500.jsonl} (500 images). Field names appearing in the remainder of this section mirror the keys in \texttt{Overall.json}. No test split is reported here; we encourage future work to reserve additional data for blinded evaluation. We summarize the per-dataset frame counts in Table~\ref{tab:split} and visualize attribute/severity coverage in Figures~\ref{fig:dataset-attributes}--\ref{fig:dataset-severity}.

\begin{table}[t]
    \centering
    \caption{Data composition by source. The validation set is disjoint from training and mirrors the geographic/climatic diversity.}
    \label{tab:split}
    \begin{tabular}{lcc}
        \toprule
        Source & Train & Val \\
        \midrule
        Argoverse1 & 754 & 80 \\
        Cityscapes & 538 & 57 \\
        KITTI & 677 & 72 \\
        nuScenes & 2\,723 & 291 \\
        \midrule
        Total & 4\,692 & 500 \\
        \bottomrule
    \end{tabular}
\end{table}

Figure~\ref{fig:dataset-attributes} summarizes the \emph{marginal distributions} of frequently used keys (e.g., \texttt{TimeOfDay}, \texttt{Weather}, \texttt{RoadType}, \texttt{Vehicles.TotalNumber}, \texttt{TrafficLights.State}). We additionally report the prevalence of \emph{list-valued} attributes (e.g., \texttt{Vehicles.Types}, \texttt{Visibility.SpecificImpairments}) via micro-averaged support counts.

\subsection{Severity (1--10) Distribution}
\label{sec:severity_dist}

Each frame includes a discrete \texttt{Severity} score in \([1,10]\). Figure~\ref{fig:dataset-severity} shows the histogram over the full dataset and the train/val splits. We also report summary statistics (median, interquartile range) to characterize the concentration of difficulty and to aid stratified sampling. 

\subsection{Artifacts and Versioning}
We release:
\begin{itemize}
  \item \textbf{JSONL annotations} with stable key names and canonical enums,
  \item a machine-readable \textbf{schema file} and a \textbf{validator},
  \item the full \textbf{prompt text} and post-processing rules,
  \item \textbf{checksums} (SHA256) for all files, plus \texttt{schema\_version}, \texttt{prompt\_hash}, \texttt{postproc\_hash}.
\end{itemize}

\subsection{Limitations}
Labels are single-frame; certain behaviors (e.g., nuanced maneuvers) can be ambiguous without temporal context. \texttt{Severity} is a coarse, scene-level proxy and does not model downstream risk under a specific ego policy. VLM outputs can reflect pretraining biases; our validation and human review mitigate but do not eliminate such effects.

\subsection{Coverage Notes and Imbalance}
\label{sec:imbalance}

We highlight coverage and potential imbalance along key axes that commonly affect training and evaluation:
\begin{itemize}
  \item \textbf{TimeOfDay \& Weather:} distributions reveal daylight dominance in some sources; we recommend stratified sampling for fair comparisons.
  \item \textbf{Traffic control prevalence:} \texttt{TrafficLights.State} appears only where lights are present; all list-field denominators exclude images without that annotation.
  \item \textbf{VRU presence:} the frequency of \texttt{Pedestrians}/\texttt{Cyclists} varies by source and geography; targeted sampling can improve robustness.
  \item \textbf{Severity spread:} mid-range severities are most common; tails (1 and 10) are naturally rarer and useful for stress tests.
\end{itemize}

\subsection{Reproducible Statistics}
\label{sec:repro_stats}

We provide scripts to regenerate Table~\ref{tab:split} and Figures~\ref{fig:dataset-attributes}--\ref{fig:dataset-severity} from the released JSONL files, ensuring that users can recompute all statistics under future schema versions. Outputs include CSV summaries and figure assets in \texttt{paper/figures/}.

\subsection{Licensing}
CARScenes combines our original materials (annotations, schema, prompt templates, validator, and code) with images from prior datasets (Cityscapes, KITTI, Argoverse v1, nuScenes). We license our annotations/schema/prompts under CC BY 4.0 and our code under MIT; these licenses apply only to our contributions. Images remain under their original licenses and are not re-licensed by us. We do not redistribute Cityscapes or nuScenes images; we provide annotations keyed to official filenames and require users to obtain those images from the original portals under their Terms of Use (nuScenes commercial use requires a separate license from Motional). KITTI and Argoverse v1 images may be redistributed only under their upstream CC BY-NC-SA licenses (KITTI: 3.0; Argoverse v1: 4.0) with attribution; if you include these images, the Non-Commercial + ShareAlike terms flow down to your bundle—otherwise release annotations-only. Consequently, any archive that includes KITTI/Argoverse imagery is non-commercial; for commercial use, exclude those images and pair our annotations with user-downloaded copies from the originals. Please cite the original datasets alongside this work.

\section{Baseline Vision-Language Model}
\label{sec:baseline}

\begin{table*}[ht]
\centering
\caption{VLM MODEL PERFORMANCE (F1) AND INFERENCE TIME (s) AGAINST HUMAN-PROVIDED GROUND TRUTH.}
\begin{adjustbox}{max width=\textwidth}
\begin{tabular}{lccccccccccccccccc}
\toprule
\textbf{} & \textbf{Scene} & \textbf{Time} & \textbf{Weather} & \textbf{Road} & \textbf{Lane} & \textbf{Signs} & \textbf{Vehicles} & \textbf{Peds}  & \textbf{Dir.} & \textbf{Ego} & \textbf{Vis.} & \textbf{Camera} & \textbf{Severity} & \textbf{Processing Time}\\
\midrule
\textbf{ChatGPT-4o} & 
0.98 & 0.99 & 0.95 & 0.99 &
0.95 & 0.95 & 0.99 & 0.98 & 0.93 &
0.93 & 0.99 & 0.98 & 0.99 & 45.63s\\
\hline
\textbf{Qwen2-VL-2B-INS} & 
0.71 & 0.99 & 0.74 & 0.99 & 0.75 & 0.76 & 0.67 & 0.81 & 0.88 & 0.75 & 0.96 & 0.99 & 0.40 & 44.58s\\
\bottomrule
\label{tab:llmvshuman}
\end{tabular}
\end{adjustbox}
\end{table*}

This section documents the \textbf{Qwen2-VL-2B-Instruct} baseline that accompanies the CARScenes dataset release. Our goal is to provide an immediately reproducible reference model together with the exact training, validation, and evaluation protocol. An abstract of the per-field model performance is given in Table ~\ref{tab:llmvshuman}.

\subsection{Model, Toolkit, and Hyperparameters}
We fine-tune \texttt{Qwen/Qwen2-VL-2B-Instruct} with the \textbf{ms-swift} toolkit using a LoRA~\cite{RefWorks:hu2022lora} recipe captured in \texttt{scripts/train\_qwen2vl.sh}. Table~\ref{tab:hyper} consolidates the key hyperparameters; the visual encoder remains frozen while LoRA adapters are injected into every language-model projection. Training runs on four 12~GiB 2080Ti GPUs (bf16) for three epochs, resulting in a wall-clock time of $\approx$50.4~hours (1,911 optimizer steps at $\approx$95~s/step, $\approx$202~GPU-hours).

\begin{table*}[t]
    \centering
    \caption{Baseline hyperparameters and schedule. Effective batch size refers to $\texttt{per\_device\_train\_batch\_size} \times \texttt{grad\_accum}$ across four GPUs.}
    \label{tab:hyper}
    \begin{tabular}{ll}
        \toprule
        Parameter & Value \\
        \midrule
        Base model & Qwen/Qwen2-VL-2B-Instruct \\
        Toolkit & ms-swift \\
        LoRA rank / $\alpha$ & 8 / 32 (decoder $q,k,v,o,\text{up},\text{down},\text{gate}$ proj.) \\
        LoRA dropout / bias & 0.05 / none \\
        Visual encoder & Frozen (\texttt{freeze\_vit=true}) \\
        Max length / max pixels & 4\,096 tokens / 1\,003\,520 pixels \\
        Optimizer & AdamW, lr=$1\times10^{-4}$, weight decay=$0$ \\
        LR schedule & Linear warmup (5\%) then constant \\
        Grad.\ clip & Disabled \\
        Batch size & 1 per GPU, grad.\ accumulation $=2$ (effective $=8$) \\
        Epochs / steps & 3 epochs, 1\,911 optimizer steps \\
        Precision & bfloat16 (AMP), decoding $T=0.0$, $p=1.0$, $\texttt{max\_new\_tokens}=512$ \\
        Validation split & Fixed file (500 images; path above) \\
        Seed & 42 (ms-swift default) \\
        Selection rule & Lowest validation loss (fixed 500-image set) \\
        Run directory & \texttt{output/qwen2vl\_lora/v3-20251009-143039} \\
        \bottomrule
    \end{tabular}
\end{table*}

\paragraph{Launch commands.} The helper script exposes all core hyperparameters through environment variables, allowing the baseline to be reproduced with:
\begin{lstlisting}[language=bash]
CUDA_VISIBLE_DEVICES=0,1,2,3 \
MAX_PIXELS=1003520 \
MAX_LENGTH=4096 \
GRADIENT_ACCUMULATION_STEPS=2 \
scripts/train_qwen2vl.sh
\end{lstlisting}
After training, deterministic inference on a folder of frames is available through:
\begin{lstlisting}[language=bash]
python scripts/infer_qwen2vl_image.py
\end{lstlisting}
which reads the prompt template from \texttt{configs/qwen\_dataset.yaml} and writes per-image predictions to \texttt{inference\_output/}. The evaluation script that produces all numbers in this section can be invoked via:
\begin{lstlisting}[language=bash]
python scripts/evaluate_validation_set.py \
  --val-file output/qwen2vl_lora/v3-20251009-143039/val_dataset_500.jsonl \
  --limit None --batch-size 8 --max-new-tokens 512
\end{lstlisting}
creating \texttt{structured\_eval\_summary.json} together with the figures in \texttt{paper/figures/}.

\subsection{Task Definition and Metrics}
The evaluation protocol treats each field in \texttt{Overall.json} as a prediction target. For categorical scalars (e.g., \texttt{Scene}, \texttt{Directionality}) we report top-1 accuracy over the exact class labels provided in the schema. List-valued attributes (e.g., \texttt{Vehicles.InMotion}) are scored with micro-averaged precision, recall, and F1 on the set difference between predicted and ground-truth elements; only images where a field is annotated contribute to the denominator. The discrete severity label ($1$--$10$) is evaluated both as a classification task (exact match) and as regression using mean absolute error (MAE) and root mean squared error (RMSE). Unless stated otherwise, model selection (Table~\ref{tab:hyper}) is based on the lowest validation loss measured on the 500-image held-out set.

\subsection{Validation Results}
\begin{figure}[t]
    \centering
    \includegraphics[width=\linewidth]{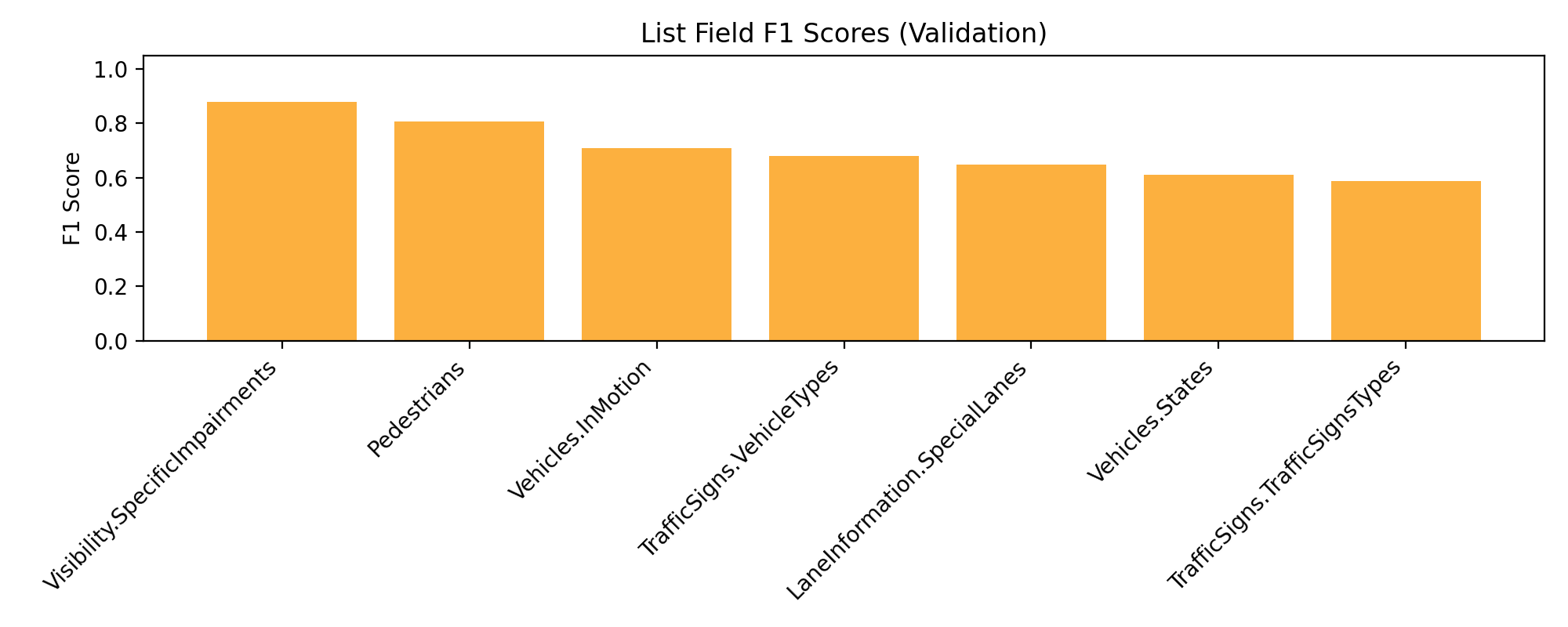}
    \caption{Micro-averaged F1 scores for list-valued attributes.}
    \label{fig:list-f1}
\end{figure}

Table~\ref{tab:scalar-acc} lists the scalar-field accuracies on the 500-image validation split (support counts reflect the number of images where the field is annotated; e.g., 162 frames contain traffic lights). Environmental cues (time, road surface, lighting) remain highly reliable ($>0.96$), while crowding levels and severity are still the dominant error sources. The corresponding list-valued metrics are shown in Table~\ref{tab:list-metrics} and Figure~\ref{fig:list-f1}, highlighting strong pedestrian detection ($\mathrm{F1}=0.82$) and visibility impairments ($\mathrm{F1}=0.89$), with moderate performance on lane/sign inventories. Severity regression attains $\mathrm{MAE}=0.72$ and $\mathrm{RMSE}=1.07$, indicating the model continues to overestimate risk by roughly one discrete level.

\begin{table}[t]
    \centering
    \caption{Scalar-field accuracy on the validation split.}
    \label{tab:scalar-acc}
    \begin{tabular}{lcc}
        \toprule
        Field & Accuracy & Support \\
        \midrule
        Scene & 0.81 & 500 \\
        TimeOfDay & 0.99 & 500 \\
        Weather & 0.82 & 500 \\
        RoadConditions & 1.00 & 500 \\
        Directionality & 0.83 & 500 \\
        Ego-Vehicle.Direction & 0.82 & 500 \\
        Ego-Vehicle.Maneuver & 0.79 & 500 \\
        Visibility.General & 0.97 & 500 \\
        CameraCondition & 0.96 & 500 \\
        TrafficSigns.TrafficSignsVisibility & 0.87 & 500 \\
        TrafficSigns.TrafficLightState & 0.85 & 162 \\
        Vehicles.TotalNumber & 0.79 & 500 \\
        Severity & 0.46 & 500 \\
        \bottomrule
    \end{tabular}
\end{table}

\begin{table}[t]
    \centering
    \caption{List-valued metrics.}
    \label{tab:list-metrics}
    \begin{tabular}{lccc}
        \toprule
        Field & Precision & Recall & F1 \\
        \midrule
        LaneInformation.SpecialLanes & 0.72 & 0.67 & 0.70 \\
        TrafficSigns.TrafficSignsTypes & 0.71 & 0.66 & 0.69 \\
        Vehicles.VehicleTypes & 0.78 & 0.72 & 0.75 \\
        Vehicles.InMotion & 0.81 & 0.81 & 0.81 \\
        Vehicles.States & 0.79 & 0.72 & 0.76 \\
        Pedestrians & 0.83 & 0.82 & 0.82 \\
        Visibility.SpecificImpairments & 0.91 & 0.87 & 0.89 \\
        \bottomrule
    \end{tabular}
\end{table}

\begin{figure}[t]
    \centering
    \includegraphics[width=\linewidth]{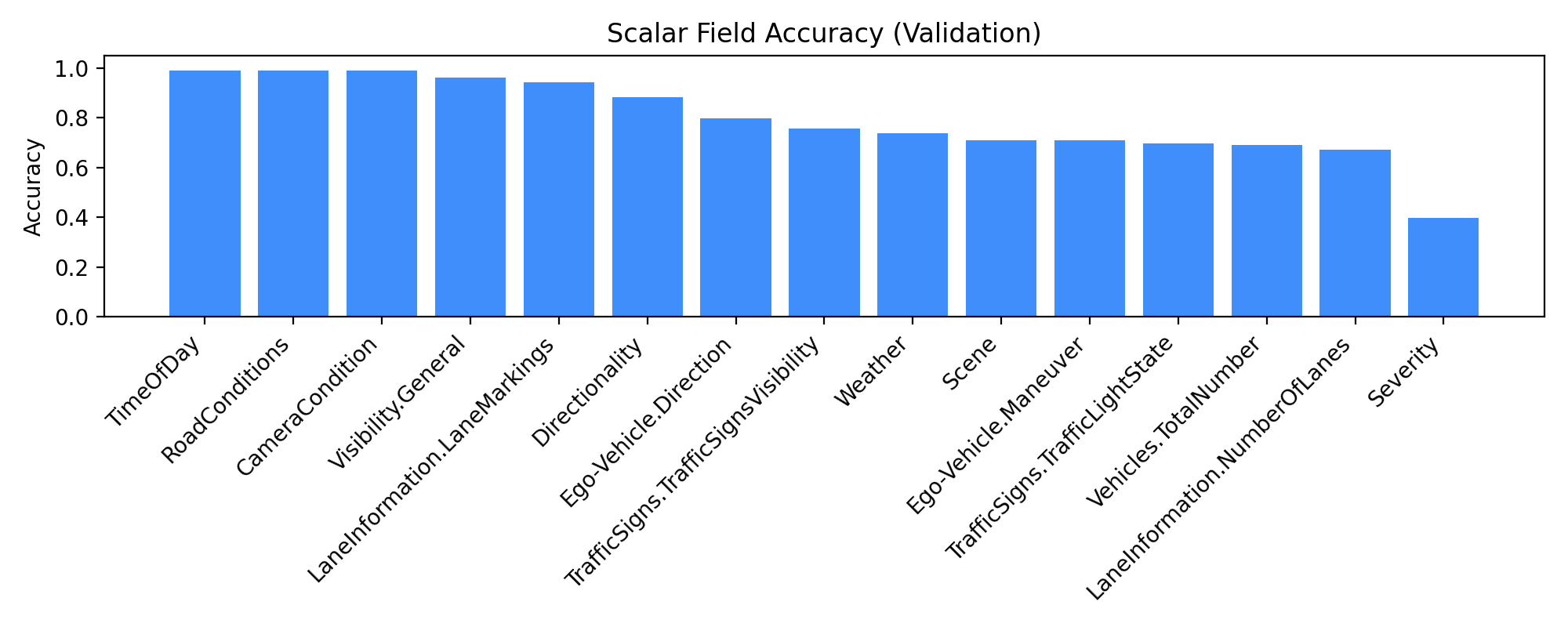}
    \caption{Scalar-field accuracy bar chart.}
    \label{fig:scalar-acc}
\end{figure}

\paragraph{Scalar-field accuracy visualization.} For completeness we include Figure~\ref{fig:scalar-acc}, the automatically generated bar chart of per-field accuracy that accompanies the release. These artifacts are committed alongside the code.

\subsection{Compute Disclosure and Reproducibility}
\begin{itemize}
    \item \textbf{Hardware:} 4$\times$NVIDIA 2080Ti GPUs (12~GiB each), bf16 compute.
    \item \textbf{Runtime:} $\approx$50.4\,h wall-clock ($\approx$202\,GPU-hours; 1,911 steps at $\approx$95\,s/step).
    \item \textbf{Software stack:} Python~3.10, ms-swift, repository commit, evaluation logs stored in \texttt{eval\_output/native/20251013\_121758}.
    \item \textbf{Artifacts:} Trained LoRA adapters reside in \texttt{checkpoint-1911}; evaluation metrics and confusion matrices are generated via \texttt{scripts/evaluate\_validation\_set.py} and \texttt{scripts/plot\_\*.py}. The resulting summary (\texttt{structured\_eval\_summary.json}, is the source of all reported numbers.
\end{itemize}

\paragraph{Future extensions.} The provided baseline serves as a foundation; we plan to release ablations (e.g., unfreezing the visual encoder, alternative prompt templates) and qualitative panels showing representative successes/failures corresponding to the confusion analysis. We will also host the LoRA checkpoint and a concise reproduction README to facilitate downstream work.

\section{Conclusion}
\label{sec:conclusion}

We introduced \textbf{CARScenes}, a lightweight, image-grounded semantic dataset for vision--language modeling in autonomous driving that augments 5{,}192 frames from standard benchmarks with a 28-key schema (350{+} attributes) and a discrete \texttt{Severity} label (1--10). Using a deterministic VLM pipeline with human verification, we release the schema, prompts, a validator, and a reproducible baseline intended to calibrate task difficulty (per-key accuracy/F1 and severity regression) rather than to set a leaderboard. CARScenes enables\textit{ 1)} structured-field training/evaluation, compositional generalization over list-valued attributes (e.g., VRUs \& night \& flashing signal), \textit{2)} coverage auditing and curation across time-of-day, weather, and infrastructure, long-tail retrieval and triage via schema queries, cross-dataset transfer under a unified ontology, curriculum/active learning using severity as a difficulty signal, and \textit{3)} scenario/test selection by mapping schema constraints to downstream simulation or closed-loop evaluations---all without requiring video or simulators.


\bibliographystyle{IEEEtran}
\bibliography{references}

@String(ECCV= {Eur. Conf. Comput. Vis.})

@String(ICLR = {Int. Conf. Learn. Represent.})

@String(AAAI = {AAAI})

@String(ECCV  = {ECCV})

@String(ICLR  = {ICLR})

@inproceedings{RefWorks:lu2023deepscenario,
	author={Chengjie Lu and Tao Yue and Shaukat Ali},
	year={2023},
	title={Deepscenario: An open driving scenario dataset for autonomous driving system testing},
	booktitle={{2023} {IEEE}/{ACM} 20th International Conference on Mining Software Repositories ({MSR})},
	publisher={IEEE},
	pages={52-56}
}

@inproceedings{RefWorks:alletto2016dr,
	author={Stefano Alletto and Andrea Palazzi and Francesco Solera and Simone Calderara and Rita Cucchiara},
	year={2016},
	title={Dr (eye) ve: a dataset for attention-based tasks with applications to autonomous and assisted driving},
	booktitle={Proceedings of the ieee conference on computer vision and pattern recognition workshops},
	pages={54-60}
}

@inproceedings{RefWorks:kung2024riskbench,
	author={Chi-Hsi Kung and Chieh-Chi Yang and Pang-Yuan Pao and Shu-Wei Lu and Pin-Lun Chen and Hsin-Cheng Lu and Yi-Ting Chen},
	year={2024},
	title={Riskbench: A scenario-based benchmark for risk identification},
	booktitle={{2024} {IEEE} International Conference on Robotics and Automation ({ICRA})},
	publisher={IEEE},
	pages={14800-14807}
}

@inproceedings{RefWorks:fremont2019scenic,
	author={Daniel J. Fremont and Tommaso Dreossi and Shromona Ghosh and Xiangyu Yue and Alberto L. Sangiovanni-Vincentelli and Sanjit A. Seshia},
	year={2019},
	title={Scenic: a language for scenario specification and scene generation},
	booktitle={Proceedings of the 40th {ACM} {SIGPLAN} conference on programming language design and implementation},
	pages={63-78}
}

@inproceedings{RefWorks:antol2015vqa,
	author={Stanislaw Antol and Aishwarya Agrawal and Jiasen Lu and Margaret Mitchell and Dhruv Batra and C. Lawrence Zitnick and Devi Parikh},
	year={2015},
	title={Vqa: Visual question answering},
	booktitle={Proceedings of the {IEEE} international conference on computer vision},
	pages={2425-2433}
}

@inproceedings{RefWorks:hudson2019gqa,
	author={Drew A. Hudson and Christopher D. Manning},
	year={2019},
	title={Gqa: A new dataset for real-world visual reasoning and compositional question answering},
	booktitle={Proceedings of the {IEEE}/{CVF} conference on computer vision and pattern recognition},
	pages={6700-6709}
}

@inproceedings{RefWorks:chen2025revisiting,
	author={Jierun Chen and Fangyun Wei and Jinjing Zhao and Sizhe Song and Bohuai Wu and Zhuoxuan Peng and S-H Gary Chan and Hongyang Zhang},
	year={2025},
	title={Revisiting referring expression comprehension evaluation in the era of large multimodal models},
	booktitle={Proceedings of the Computer Vision and Pattern Recognition Conference},
	pages={513-524}
}

@article{RefWorks:krishna2017visual,
	author={Ranjay Krishna and Yuke Zhu and Oliver Groth and Justin Johnson and Kenji Hata and Joshua Kravitz and Stephanie Chen and Yannis Kalantidis and Li-Jia Li and David A. Shamma},
	year={2017},
	title={Visual genome: Connecting language and vision using crowdsourced dense image annotations},
	journal={International journal of computer vision},
	volume={123},
	number={1},
	pages={32-73}
}

@article{RefWorks:hu2022lora,
	author={Edward J. Hu and Yelong Shen and Phillip Wallis and Zeyuan Allen-Zhu and Yuanzhi Li and Shean Wang and Lu Wang and Weizhu Chen},
	year={2022},
	title={Lora: Low-rank adaptation of large language models.},
	journal={ICLR},
	volume={1},
	number={2},
	pages={3}
}

@inproceedings{RefWorks:kim2018textual,
	author={Jinkyu Kim and Anna Rohrbach and Trevor Darrell and John Canny and Zeynep Akata},
	year={2018},
	title={Textual explanations for self-driving vehicles},
	booktitle={Proceedings of the European conference on computer vision ({ECCV})},
	pages={563-578}
}

@inproceedings{RefWorks:ding2024holistic,
	author={Xinpeng Ding and Jianhua Han and Hang Xu and Xiaodan Liang and Wei Zhang and Xiaomeng Li},
	year={2024},
	title={Holistic autonomous driving understanding by bird's-eye-view injected multi-modal large models},
	booktitle={Proceedings of the {IEEE}/{CVF} Conference on Computer Vision and Pattern Recognition},
	pages={13668-13677}
}

@inproceedings{RefWorks:qian2024nuscenesqa,
	author={Tianwen Qian and Jingjing Chen and Linhai Zhuo and Yang Jiao and Yu-Gang Jiang},
	year={2024},
	title={Nuscenes-qa: A multi-modal visual question answering benchmark for autonomous driving scenario},
	booktitle={Proceedings of the {AAAI} Conference on Artificial Intelligence},
	volume={38},
	chapter={5},
	pages={4542-4550}
}

@article{RefWorks:wang2024qwen2vl,
	author={Peng Wang and Shuai Bai and Sinan Tan and Shijie Wang and Zhihao Fan and Jinze Bai and Keqin Chen and Xuejing Liu and Jialin Wang and Wenbin Ge},
	year={2024},
	title={Qwen2-vl: Enhancing vision-language model's perception of the world at any resolution},
	journal={arXiv preprint arXiv:2409.12191}
}

@article{RefWorks:liu2023visual,
	author={Haotian Liu and Chunyuan Li and Qingyang Wu and Yong Jae Lee},
	year={2023},
	title={Visual instruction tuning},
	journal={Advances in neural information processing systems},
	volume={36},
	pages={34892-34916}
}

@inproceedings{RefWorks:sima2024drivelm,
	author={Chonghao Sima and Katrin Renz and Kashyap Chitta and Li Chen and Hanxue Zhang and Chengen Xie and Jens Beißwenger and Ping Luo and Andreas Geiger and Hongyang Li},
	year={2024},
	title={Drivelm: Driving with graph visual question answering},
	booktitle={European conference on computer vision},
	publisher={Springer},
	pages={256-274}
}

@article{RefWorks:deruyttere2019talk2car,
	author={Thierry Deruyttere and Simon Vandenhende and Dusan Grujicic and Luc Van Gool and Marie-Francine Moens},
	year={2019},
	title={Talk2car: Taking control of your self-driving car},
	journal={arXiv preprint arXiv:1909.10838}
}

@inproceedings{RefWorks:yu2020bdd100k,
	author={Fisher Yu and Haofeng Chen and Xin Wang and Wenqi Xian and Yingying Chen and Fangchen Liu and Vashisht Madhavan and Trevor Darrell},
	year={2020},
	title={Bdd100k: A diverse driving dataset for heterogeneous multitask learning},
	booktitle={Proceedings of the {IEEE}/{CVF} conference on computer vision and pattern recognition},
	pages={2636-2645}
}

@inproceedings{RefWorks:sun2020scalability,
	author={Pei Sun and Henrik Kretzschmar and Xerxes Dotiwalla and Aurelien Chouard and Vijaysai Patnaik and Paul Tsui and James Guo and Yin Zhou and Yuning Chai and Benjamin Caine},
	year={2020},
	title={Scalability in perception for autonomous driving: Waymo open dataset},
	booktitle={Proceedings of the {IEEE}/{CVF} conference on computer vision and pattern recognition},
	pages={2446-2454}
}

@inproceedings{RefWorks:chang2019argoverse,
	author={Ming-Fang Chang and John Lambert and Patsorn Sangkloy and Jagjeet Singh and Slawomir Bak and Andrew Hartnett and De Wang and Peter Carr and Simon Lucey and Deva Ramanan},
	year={2019},
	title={Argoverse: 3d tracking and forecasting with rich maps},
	booktitle={Proceedings of the {IEEE}/{CVF} conference on computer vision and pattern recognition},
	pages={8748-8757}
}

@inproceedings{RefWorks:caesar2020nuscenes,
	author={Holger Caesar and Varun Bankiti and Alex H. Lang and Sourabh Vora and Venice Erin Liong and Qiang Xu and Anush Krishnan and Yu Pan and Giancarlo Baldan and Oscar Beijbom},
	year={2020},
	title={nuscenes: A multimodal dataset for autonomous driving},
	booktitle={Proceedings of the {IEEE}/{CVF} conference on computer vision and pattern recognition},
	pages={11621-11631}
}

@inproceedings{RefWorks:cordts2016cityscapes,
	author={Marius Cordts and Mohamed Omran and Sebastian Ramos and Timo Rehfeld and Markus Enzweiler and Rodrigo Benenson and Uwe Franke and Stefan Roth and Bernt Schiele},
	year={2016},
	title={The cityscapes dataset for semantic urban scene understanding},
	booktitle={Proceedings of the {IEEE} conference on computer vision and pattern recognition},
	pages={3213-3223}
}

@article{RefWorks:geiger2013vision,
	author={Andreas Geiger and Philip Lenz and Christoph Stiller and Raquel Urtasun},
	year={2013},
	title={Vision meets robotics: The kitti dataset},
	journal={The international journal of robotics research},
	volume={32},
	number={11},
	pages={1231-1237}
}

\end{document}